\documentclass{article} 
\usepackage{11785_project,times}
\usepackage[hidelinks]{hyperref}
\usepackage{url}
\usepackage{amsmath}
\usepackage{graphicx}
\usepackage{float}
\usepackage{animate}

\title{Activity Recognition on a Large Scale in Short Videos - Moments in Time Dataset}

\author{Ankit Parag Shah* \thanks{First four authors contributed equally} \\
\texttt{\small aps1@andrew.cmu.edu} 
\And
 \textbf{Harini Kesavamoorthy*} \\
\texttt{\small hkesavam@andrew.cmu.edu} 
\And
\textbf{Poorva Rane*} \\
\texttt{\small prane@andrew.cmu.edu} 
\And
\textbf{Pramati Kalwad*} \\
\texttt{\small pkalwad@andrew.cmu.edu} 
\And 
\textbf{Alexander Hauptmann} \\
\texttt{\small alex@cs.cmu.edu}
\And
\textbf{Florian Metze}\\
\texttt{\small florian@cs.cmu.edu}
}

\begin{document}
\maketitle
\begin{abstract}
Moments capture a huge part of our lives. Accurate recognition of these moments is challenging due to the diverse and complex interpretation of the moments. Action recognition refers to the act of classifying the desired action/activity present in a given video.  In this work, we perform experiments on Moments in Time dataset to recognize accurately activities occurring in 3 second clips. We use state of the art techniques for visual, auditory and spatio temporal localization and develop method to accurately classify the activity in the Moments in Time dataset. Our novel approach of using Visual Based Textual features and fusion techniques performs well providing an overall 89.23 \% Top - 5  accuracy on the 20 classes - a significant improvement over the Baseline TRN model. 

\end{abstract}

\section{Introduction}
Modeling the spatial-audio-temporal dynamics for actions in videos poses multiple challenges since meaningful events may include objects, animals and natural phenomena in addition to people. Moreover, visual and auditory events may be transient or sustained, and can be symmetrical or asymmetrical. In this project, we aim to solve the problem of human concept understanding in very short videos. We attempt to solve this problem for the Moments in Time Dataset which is decribed in detail in Section \ref{dataset}. This dataset is unique as it contains everyday actions which may have several interpretable forms, such as `opening' which is a vague verb that could indicate opening of a gift or opening of eyes. This makes the dataset challenging and a very interesting problem to work on.

Action Recognition is a well-explored problem with several challenges and papers on the same. Some of the popular action recognition tasks are ActivityNet Recognition(\cite{caba2015activitynet}) challenges which contain tasks on Human Action Recognition, as well as Sports Action recognition. The problem chosen for this paper not only contains actions, but contains these actions in every interpretable form. This leads to very different approaches of solving the problem from conventional Human Action Recognition tasks. 

We explored different modalities for extracting features for final classification. In the visual domain, we consider Spatial features and Spatio-Temporal features. ResNet-152 (\cite{7780459}) has been the state-of-the-art model for performing the task of image classification. We use ResNet 152 which is an improvement over ResNet-50 as described in the paper (\cite{7780459}). Audio contains complementary information as compared with Visual Features and there is less exploration mentioned in the Moments in Time Paper (\cite{monfortmoments}). Hence we use better feature representation over the SoundNet features as described in (\cite{kumar2017knowledge}) with a modification necessary for Moments in Time dataset. In the Spatio-Temporal modality, we explore the 3D convolution based architectures. The newly proposed ResNext architecture \cite{xie2017aggregated} compares the performance of ResNet152, ResNet101 architecture and outperforms these models. Hence, we explore the ResNext architecture with 3D convolutions in order to capture the spatio-temporal relations in the video. We discuss the limitations of this model for the moments in time dataset, and also propose a novel method of overcoming these limitations. We propose `VisText', a method of extracting text features from a videos and also discuss its advantages/disadvantages in this report. To the best of our knowledge, this is the first time text modality has been introduced into video classification task for datasets that do not contain text input. We also perform the fusion of these modalities and present the results. There is a possibility of having more than one action taking place in a three second long video and action recognition models that correctly predict one of these actions may be wrongly penalized because the ground truth does not include that action. To avoid hindrance in their performance, we employ the top-1 and top-5 accuracy measure to report classification performances. 

\section{Dataset}
\label{dataset}
\subsection{Moments in Time Dataset}
A collection of one million videos (802,264 training videos, 33,900 validation videos and 67,800 testing videos) each 3 seconds long with one action label per video(\cite{monfortmoments}). The videos are pulled from 10 different sources. The videos in the collection capture visual and/or audible short events, produced by humans, animals, objects or nature. The dataset comprises of 339 different classes and the classes are chosen such that they include the most commonly used verbs in the English language, covering a wide and diverse semantic space. The two key features of the Moments in Time dataset are diversity and scale. Another feature of the Moments in Time dataset is that there are sound-dependent classes included. We evaluate our methods on Moments in Time Track - 20 class dataset and then use the insights gained to perform experiments for Moments in Time Mini Track - 200 classes dataset. 

\subsubsection{Challenges}
The dataset contains videos of not just plain actions or objects generally involved in an action (eg: balls in juggling), but the underlying concept within each action. This makes the MIT Dataset very challenging for conventional methods like extracting spatio-temporal relations to derive meaningful representations of the data. The figure below shows the video snapshots with the labels. As we can see, there is a certain `common-sense' needed to learn from these videos.
\begin{figure}[H]
\begin{center}
\includegraphics[width=\linewidth]{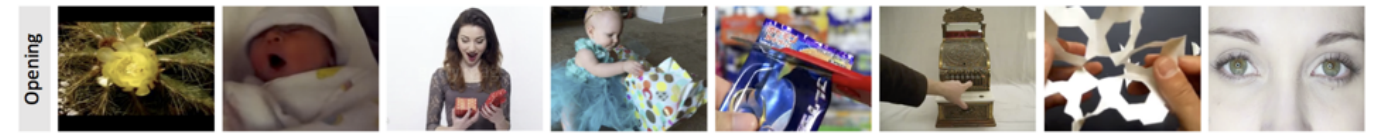}
\includegraphics[width=\linewidth]{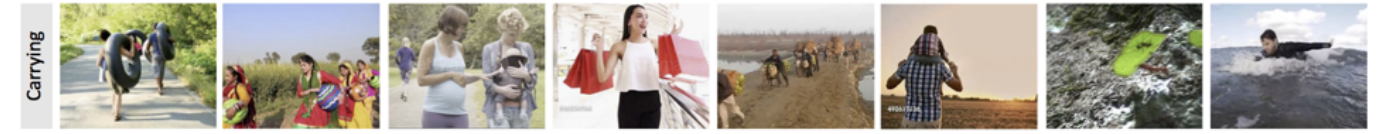}
\end{center}
\end{figure}

\section{Related Work}

Videos unlike images contain a wealth of information. The practicality of video have led to advancement for video recording, viewing and distribution. However, given the amount of video data available it is impossible to extract useful information manually at a large scale.Thus, there is a need for developing automatic techniques which can efficiently locate interesting parts containing information within a video stream. Being able to detect actions within a video stream provides useful information which would be otherwise difficult to manually annotate. 

Traditional approaches in activity recognition deals were performed on untrimmed videos. Thus, being able to localize the action in the video stream is important. The candidate temporal or spatio-temporal regions are referred as action proposals. Action proposals don't provide information about the class of an action but provides localize regions which are most likely to contain an action as mentioned in this extensive review \cite{kang2016review} paper . Action proposals (or non-action shots) prevent going through an exhaustive search space and helps detect temporal or spatio-temporal locations of actions.

Paper \cite{xiong2016cuhk} describes that the models trained on Kinetics dataset serves as a better pre-trained model as compared with the ImageNet model. The incorporate temporal information using 3D models and long temporal information using temporal segment networks. Further, their approach also explored multi-modal cues such as audio cues and body pose estimation. Paper \cite{feichtenhofer2016convolutional} mentions 

In \cite{venugopalan2015sequence} optical flow features between consecutive frames are used to model temporal aspects of activities depicted in the videos. It was shown by \cite{ng2015beyond} optical flow features capture crucial information for the task of activity classification. \cite{venugopalan2015sequence} use the features obtained in the fc6 layer activations in the CNN pre-trained on the UCF101 video dataset to represent the activity for the task of activity classification. Deep convolutional networks can also be employed to capture the complementary information contained in still frames and motion between frames. In order to incorporate spatial and temporal features \cite{simonyan2014two} propose a two-stream ConvNet architecture and demonstrate that a ConvNet trained on multi-frame dense optical flow is able to achieve very good performance despite having limited training data. They also we show that applying multitask learning to two different action classification datasets can be used to increase the amount of training data and improve the performance on both. UCF101 dataset and Thumos dataset built from web videos have become benchmarks for video classification. 

Moments in Time Dataset \cite{monfortmoments} has video information of 3 second clips. Temporal events of 3 seconds corresponds to average duration of human working memory. Bundling 3 seconds of actions together will allow to create complex long sequence of actions such as picking up an object and carrying the object to another place can be broken into picking, carrying and such modular actions. Action refers to the motion of the human body which may or may not be cyclic. 

Challenge with Moments in Time Dataset task is to develop models which can understand the different transformations in a way to allow them to discriminate between different actions and still generalize the task of action recognition for longer activities. 

Most of the previous methods transfer the image level object recognition representation to the video level action recognition representation. They argue that a natural approach is to learn general level of video level representation and then apply it to other video analysis applications. 

\section{Baseline}
The baseline paper (Moments in Time) \cite{monfortmoments} employs Temporal Relation Networks. (TRN-Multiscale)\cite{zhou2017temporal}. This paper proposes a unique method combining the spatial relations with temporal relation by randomly sampling frames from the video in a sorted fashion, extracting their Spatial features from Convolutional layers. These features are then taken with 2 frame relation, 3 frame relation, 8 frame relations which are passed through an MLP. The frames are randomly sampled and in a specific order where 1,9th frame is an allowed order but not 9,1 pair in the 2 frame relation. These new relations are aggregated together using an addition operation. These features are then classified into one of the moments classes. 

\begin{figure}[]
\begin{center}
\includegraphics[scale=0.4]{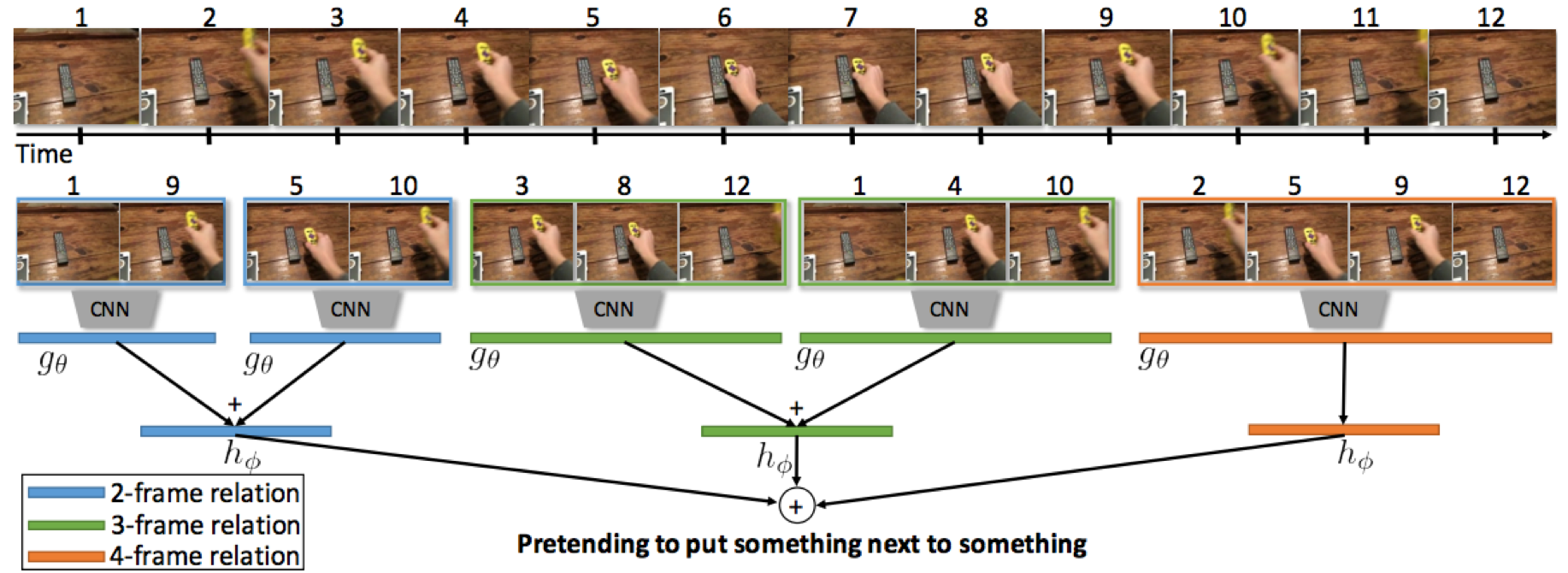}
\caption{Baseline architecture}
\label{fig:baselinearch}
\end{center}
\end{figure}

\section{Methodology}
We use three different modalities for the purpose of classification.
\begin{itemize}
\item Audio Features
\subitem Log Mel Spectrogram
\subitem SoundNet conv 4 and conv 5 layer features
\subitem Weakly Labelled Audio (WALNet) feature representation at the segment level
\item Visual Features
\subitem Spatial Features
\subitem Spatio-Temporal Features
\item Text Features
\subitem VisText Feature
\end{itemize}

\subsection{Audio Features}

Moments in time dataset has 3 second clips with audio files in them. We ran a quick analysis to find out the percentage of files containing the audio for the training set. We found 56\% [56014 out of 100000] of the files in the training set contains the audio file with the visual feed. However, the validation set from Moments in time dataset contains 62.8\% [6287 of 10000] of the files containing the audio feed which we believe will be close to the test set whose details will be released in the month of April. \\
\subsubsection{Log Mel Spectrogram}
We extract the log mel spectrogram feature representation since log mel spectrograms have been proven to work well over mfcc features for audio event classification tasks in general. For log mel spectrogram extraction, we use Librosa library 0.6.1 version and initially resample the audio at 44.1kHz prior to feature extraction. We use hop size as 512 with a window length of 1024 and extract 128 mel filters for the audio recording. 

\subsubsection{SoundNet Feature}
SoundNet \cite{aytar2016soundnet} uses raw waveforms to train the network by transferring discriminative knowledge from the visual recognition networks into sound networks. Their 8-layer architecture achieves 72.9 percent accuracy on the ESC-50 dataset which was the state of the art result in 2016. Moments in Time Dataset \cite{monfortmoments} shows that the auditory model quantitatively performed different as compared with the visual model suggesting that the sound has complementary information as compared with vision for recognition of action in videos. Thus, we believe that our effort in extracting this audio feature would enable to learn better representation for audio as modality over log mel spectrograms as well. We extract conv 4 and conv 5 layers for feature extraction and use it during further audio event based classification\\
\subsubsection{WALNet architecture feature representation}
\begin{figure}[h]\centering
\includegraphics[width=0.9\linewidth]{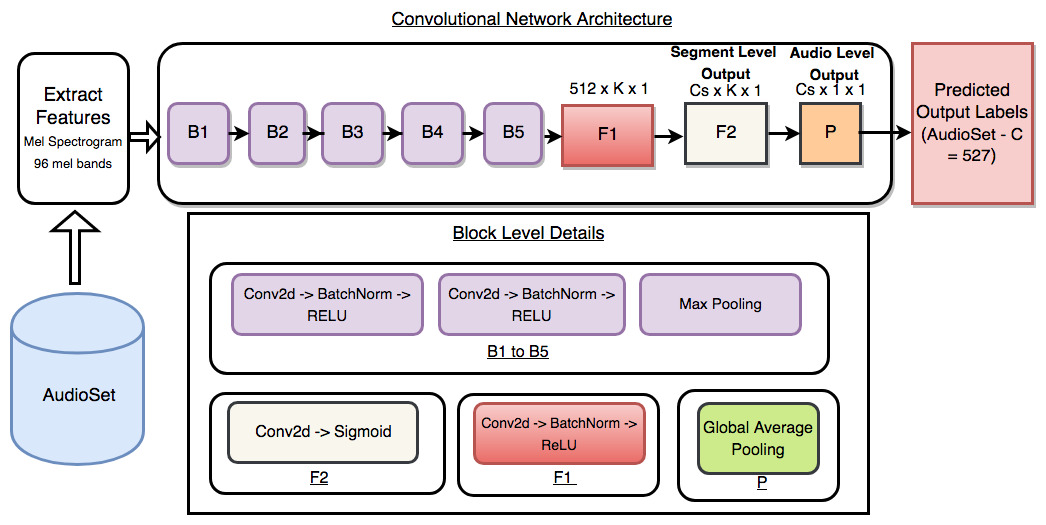}
\caption{WALNet architecture for feature extraction pipeline. The layer blocks from L1 to L6 consists of 2 or 1 convolutional layers followed by max pooling. L8 represents segment level output which is further mapped by the global average pooling layer P to produce recording level output. The loss is then computed with respect to the target recording level labels. }
\label{fig:audio_feature}
\end{figure}

Architecture as described in \cite{kumar2017knowledge,shah2018closer} improves over SoundNet audio feature representation using the architecture as described in figure \ref{fig:audio_feature}. For the feature extraction we will use log scale mel bands as feature and convert the feature representation using successive dimension reduction and batch normalization with B1 - B5 layers and passing through fully connected layer to represent the audio file at a segment level. We assume that the label corresponding to each segment of the audio file represents the label for the entire video recording.

Architecture in Figure \ref{fig:audio_feature} requires the input to be of a constant length but can handle variable length audio recordings and hence we divide the data into multiple batches with appropriate sizes [determined based on the training data]. On empirical analysis of the dataset, we find that the dataset doesn't contain all video files with 3 second recordings and thus, design choices such as appending mean as the final layer output or zero padding and other padding solutions needs to be explored for this dataset - which will have a minor impact given the duration range is 2.87 seconds to 3.05 seconds. 

For WALNet based features for Moments in Time, [\ref{fig:audio_feature}] we implement the feature extraction pipeline.  The blocks of layers L1 to L5 consists of two convolutional layers followed by a max pooling layer. The convolutional layers consists of batch normalization before non-linear activation function. ReLU ($max(0,x)$) \cite{nair2010rectified} is used in all layers from L1 to L6. $3 \times 3$  convolutional filters are used in all layers from L1 to L6. Stride and padding values are fixed to $1$. The number of filters employed in different layers is as follows, \emph{\{L1: 16, L2:32, L3:64, L4:128, L5:256, L6:512 \}}. Note that L1 to L6 consists of two convolutional layers followed by max pooling. Both convolutional layers in these blocks employ same number of filters. The max pooling operation is applied over a window of size $2 \times 2$. Logmel inputs are treated as single channel inputs. The convolutional layers in L1 does not change the height and width of input. The max pooling operation reduces the size by a factor of $2$. For example, a recording consisting of $128$ logmel frames that is an input of size $X \in R^{1 \times 128 \times 128}$, will produce and output of size $16 \times 64 \times 64$ after L1.  This design aspect applies  to all layer blocks from L1 to L6 and hence $1 \times 128 \times 128$ input will produce a $512 \times 2 \times 2$ output after L6. L7 is a convolutional layer with $1024$ filters of size $2 \times 2$. Once again ReLU activation is used. Stride is again fixed to $1$ and no padding is used. 

\subsection{Visual Features}
\subsubsection{Spatial Features}
\label{SpatialFeatures}
In the base architecture, the authors experiment with finetuning a ResNet-50 architecture with three different strategies \cite{monfortmoments}. They train the network from scratch, initialised on the Places dataset and initialised on the ImageNet dataset. They sample 6 frames from the video and average the results in order to obtain the final classification scores. They conclude that the ResNet-50 architecture fine-tuned with initialisation on the ImageNet dataset produces the best results with a Top 1 Accuracy of 27.16\% and a Top 5 error of 51.68\%. The results are further explored in the Experiments and Results section.

With these learnings, we continue our experimentation with the DenseNet-161, ResNet-152 and the VGG Net-19 with Batch Normalisation, all of which have comparable results with ResNet-50 on the ImageNet dataset. The results are shown in table \ref{ImageNetVisual} . We finetune the architectures by changing the number of units in the last fully connected layer to represent the number of classes in moments (200) from the number of classes in the ImageNet dataset (1000). For testing, we pass all the frames of a video through the network and perform an average on them to obtain the final classification results. The results of these architectures are discussed in the Experiments and Results section. For the fusion techniques, we obtain the output of the bottleneck layers, just before the final fully connected layer for early fusion, and obtain the final output of the networks for late fusion. \vspace{-5mm}
\begin{table}[h]
\begin{center}
\caption{Results on ImageNet dataset}
\begin{tabular}{|c|c|c|}
\hline
Network & Top-1 Error& Top-5 Error\\ \hline
ResNet-50& 23.85\% & 7.13\%\\
ResNet-152 & 21.69\% & 5.94\%\\
DenseNet-161 & 22.35\% & 6.20\%\\
VGG-19 with batch normalization & 25.76\% & 8.15\%\\ \hline
\end{tabular}
\end{center}
\label{ImageNetVisual}
\end{table}
\subsubsection{Spatio-Temporal Features}
Very deep Convolutional Neural Networks (CNNs) with spatio-temporal 3D kernels outperform state-of-the art complex 2D architectures for activity recognition in videos as seen by \cite{hara2017can}. ResNext is one such homogeneous, multi-branch architecture that has only a few hyper-parameters to set. This introduces a new-dimension called "cardinality" in addition to the dimensions of depth and width. \cite{xie2017aggregated} show that  even under the restricted
condition of maintaining complexity, increasing cardinality
is  is more effective than going deeper or
wider to improve classification accuracy on the Imagenet-1k dataset. 

The transformation functions in this architecture extends the VGG-style strategy of repeating layers of the same shape, which is helpful for isolating a few factors
and extending to any large number of transformations. ResNext uses grouped convolutions. In this architecture, all the low-dimensional embeddings can be replaced by a single, wider layer. Splitting is essentially done by the grouped convolutional layer when it divides its input channels into
groups. The grouped convolutional layer performs groups of convolutions whose input and output channels are 4-dimensional. The grouped convolutional layer concatenates them as the outputs of the layer. As a result, the actions in a video can be represented more effectively when compared to ResNet-152 while having fewer parameters to learn. \vspace{-2mm}
\subsection{Text Features}
Moments In Time Dataset claims to be a dataset which has a common-sense interpretation of each activity, we need to extract some semantic features from the three-second videos. Our general hypothesis is that this sort of semantic relationship is abundant in text, which can be easily trained on a much larger (richer) dataset. 
\textbf{Initial Hypothesis:}
We propose to use the word embedding of the actual ImageNet labels to represent each video from Word2Vec trained on Google News Corpus. We suspect this would be way to incorporate an outside “knowledge” into our system. Furthermore, we could choose to use top-k labels for the embedding and use this embedding to learn the word embedding of the class label (like “opening”, “carrying”). This network would then be trained on all the training data, to give us the textual features, which is technically an n-dimensional vector of the learned label embedding. This label embedding is our final textual feature as input to our classifier.

\begin{figure}[H]
\includegraphics[scale=0.45]{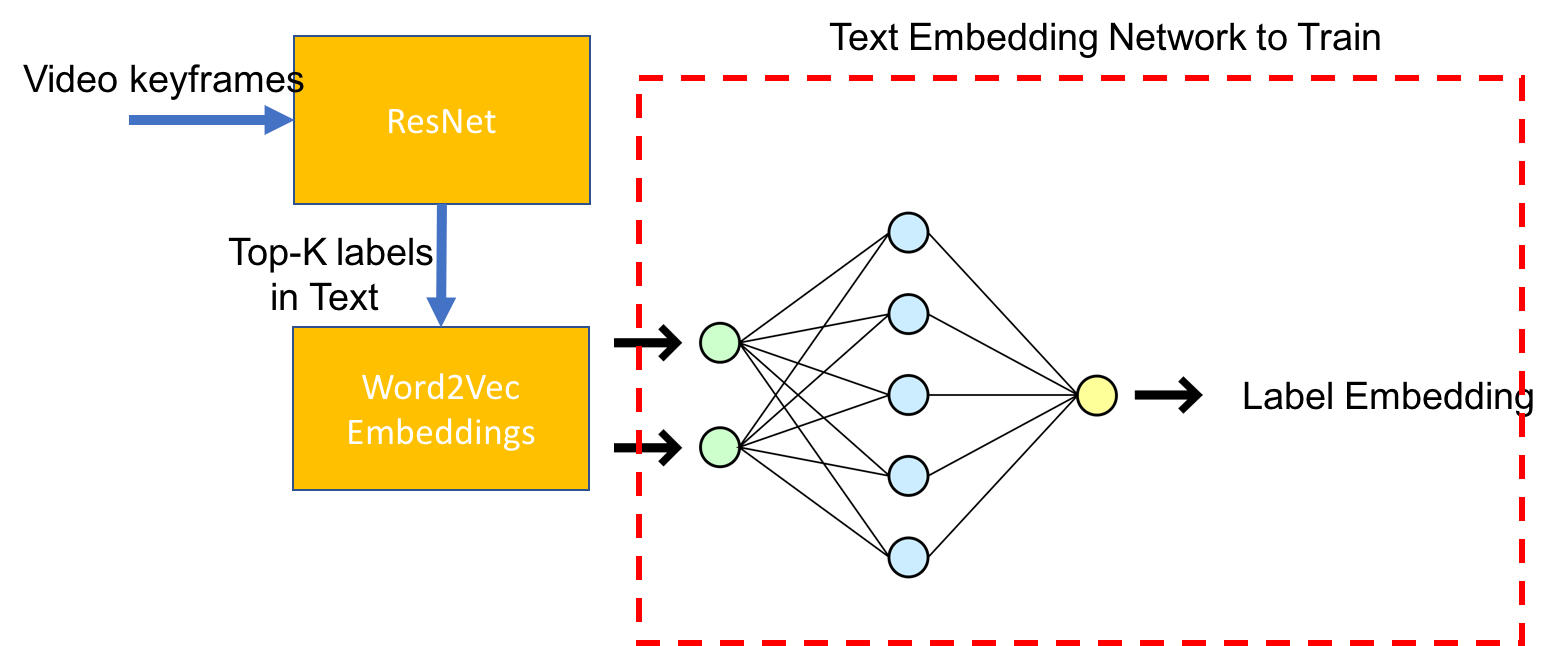}
\caption{Architecture for producing textual features.}\vspace{-5mm}
\end{figure}
We performed toy experiments with this data. Firstly, the ResNet50 output layer of dimension 1x1000 was extracted, where each element corresponds to the probability of an ImageNet class. The top five labels for this were taken and their actual word for the label was converted into it's word2vec representation. Upon examining the label words along with the Moments class label words, we find little correlation. After performing the classification using the five word2vec representation with respect to the moments classes, we obtain a classification accuracy of only ~5\%. This showed that our initial hypothesis was refuted. We suspect that since at the last layer of ResNet50 architecture, a lot of the feature representation is lost due to classification, the ImageNet labels need not be semantically coherent with the moments labels for every video. Hence, we alter our text feature generation for a better representation of the video.\\

\textbf{Final Hypothesis:} The Moments classes can be easily confused for another even by humans. We hypothesize that subtle differences in these classes like \textit{exercising} and \textit{running} can be captured better in a higher dimensional space than just a single number. Thus, the difference in the word2vec representation of these class labels which have been trained on a generic corpus, can provide more information which can help resolve the differences between these classes. We propose a novel way to extract text features from visual features called `VisText' features described in detail in section \ref{vistextsection}.
\subsubsection{VisText Feature}
\label{vistextsection}
\begin{figure}[H]
\begin{center}
\includegraphics[trim={0 0 5.3cm 0},clip,scale=0.4]{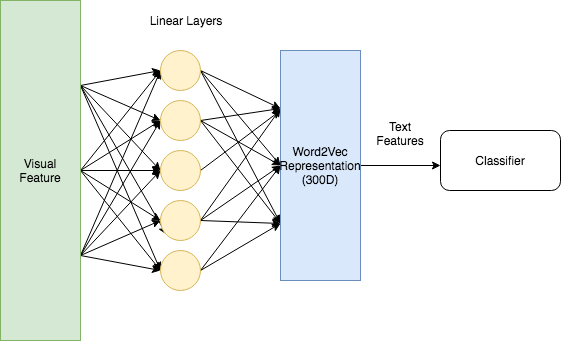}
\caption{Architecture for generating VisText features.}
\label{vistext}
\end{center}
\end{figure}

Converting vectors from one modality to another has been explored in the past for various problems. The Deep Attentional Multimodal Similarity Loss in the paper \cite{xu2017attngan} used a method of converting image and text data into a similar space by passing it through a Multi-Layer perceptron. Another attempt at normalizing modality is in the paper \cite{dong2016word2visualvec}, which aims at solving the problem of cross-modal inferences by using stacked linear layers in image captioning scenario. Here, the word2vec representation of the captions are passed through a trained model of multi-layered perceptrons to obtain a visual representation of the text. It is trained such that the target is a ResNet50 or a ResNet152 output of the image. This leads us to our problem which aims at doing the reverse. We need to obtain the word2vec representation of the visual features that are input to our model. Figure \ref{vistext} shows the high level diagram of extracting features from visual features.

In this paper we follow the following architecture for training the VisText model.
\begin{itemize}
\item Input: Visual Feature (ResNet-50,3D ResNext-101)
\item Three Layer Multi-Layer Perceptron with ReLu activations and Batch Normalization
\item Loss: Huber Loss (L1 Regularised MSELoss)
\item Target: 300d word2vec of class label word.
\item (Eval) Output: 300d word2vec resembling
\end{itemize}

This model uses an element wise Huber Loss which is given by:
\begin{equation}
    loss(x, y) = \frac{1}{n} \sum_{i} z_{i}
\end{equation}
where, $z_{i}$ is given by:
\begin{equation}
    z_{i} = \begin{cases}
    0.5 (x_{i} - y_{i})^{2} & \text{if} |x_i - y_i| < 1 \\
    |x_{i} - y_{i}| - 0.5 & \text{otherwise }
    \end{cases}
\end{equation}

We use the word2vec representation which was trained on the Google news corpus. 
We obtain results for VisText features when we conduct experiments on the final classification with different visual features which is shown in Table \ref{tab:visTextResults}.







\section{Evaluation Metrics}
We use the evaluation metrics as proposed on the Moments in Time dataset. The two evaluation metrics - top 1 accuracy and top 5 accuracy. 
Top-1 accuracy indicates the percentage of testing videos for which the top confident predicted label is correct. Top-5 accuracy indicates the percentage of the testing videos for which the ground- truth label is among the top 5 ranked predicted labels. Top-5 accuracy is appropriate for video classification as videos may contain multiple actions within them.

For each video, an algorithm will produce k labels lj, j=1,..,k. The ground truth label for the video is g. The error of the algorithm for that video would be
\begin{equation}
    e=\begin{cases} min_{jd}(l_j,g)  & \text{ with d(x,y)=0 if x=y} \\  1 & \text{otherwise} \end{cases}.
\end{equation}  
The overall error score for an algorithm is the average error over all videos.



\section{Experiments and Results}
We discuss all our experiments and results in detail in the following section. Our initial discussion is based on experiments performed on individual feature representation and then later we dive into success and failure cases with extensive ablation study performed on the dataset. Further, we perform ensemble based approaches with fusion techniques and show performance improvements using all feature representation. 
\subsection{Baseline Performance - Moments in Time}
\begin{figure}[H]
\begin{center}
\includegraphics[scale=0.3]{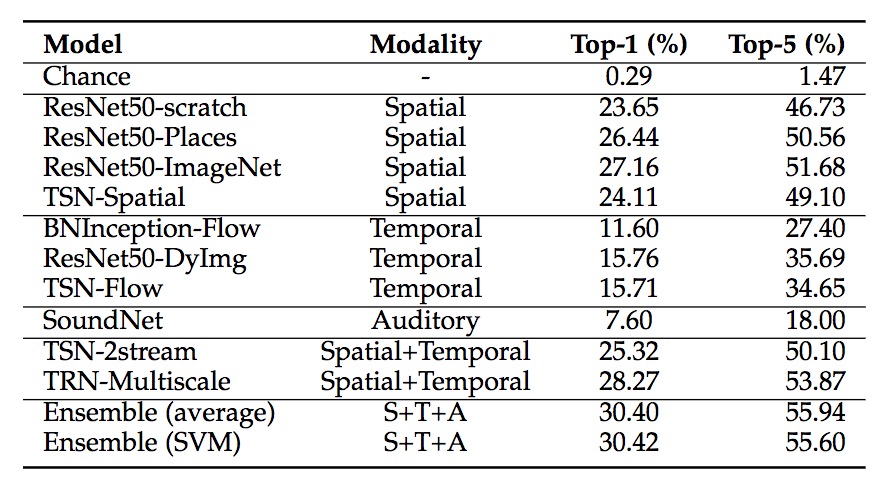}
\caption{Baseline performance on the validation set as described in Moments in Time Paper on Full dataset}
\label{fig:baseline}
\end{center}
\end{figure}

The below figure \ref{fig:baseline} represents the baseline results as obtained in the Moments in Time Dataset \cite{monfortmoments} paper. We observe that Ensemble method performs well as compared with individual features and that is justified since each modality might represent complementary information about a given activity under question. We note that the audio features - SoundNet performs about 7.6 percent which is conforming to our hypothesis since audio features will give less improvements given the number of activities present in the dataset that can be purely identified using audio features as well as the fact that roughly half the videos doesn't contain audio information. 

\subsection{Audio Features Analysis}
We observe that our model doesn't perform well using Log Mel Spectrogram which is expected behaviour since the performance of log mel spectrogram is not enough for audio event classification. Thus, there are other feature extraction techniques explored to enhance the information extracted using log mel spectrograms. \\ Further, we confirm our hypothesis that SoundNet features will perform better than log mel spectrograms given their performance on ESC-50 dataset is better as well as WALNet architecture will perform best among all the audio features extracted. Since we observe WALNet features extracted performs best among the audio features, we later use this representation in our fusion tasks as explained in Section on Fusion Techniques. 

\begin{table}[h]
\centering
\caption{Audio Event Analysis for different architecture design}\vspace{2mm}
\label{my-label}
\begin{tabular}{|l|l|}
\hline
Model                     & Top-1 Accuracy (\%) \\ \hline
Log Mel Spectrogram (SVM) & 5.6                 \\ \hline
SoundNet conv\_5 (SVM)    & 10.9                \\ \hline
WAL-Net Architecture      & 19.7                \\ \hline
\end{tabular}
\end{table}

We extract the audio feature representation for 20 classes and view the class-wise performance using the WALNet architecture feature representation. We observe that categories like hugging and reading which are very difficult to be identified using audio feature representation don't perform as well and this conforms with our hypothesis. Further, events such as crying and welding which have specific acoustic characteristics are represented accurately and identified using audio features. 

\begin{table}[H]
\centering
\caption{Audio analysis - WALNet architecture - 20 classes}\vspace{2mm}
\label{my-label}
\begin{tabular}{|l|l|l|l|}
\hline
\textbf{Class} & \textbf{Accuracy} & \textbf{Class} & \textbf{Accuracy} \\ \hline
arresting,     & 0.219             & gardening      & 0.156             \\ \hline
attacking      & 0.050             & hammering      & 0.251             \\ \hline
baking         & 0.082             & hugging        & \textbf{0.066}             \\ \hline
bulldozing     & 0.284             & juggling       & 0.220             \\ \hline
camping        & 0.125             & opening        & 0.084             \\ \hline
chewing        & 0.215             & painting       & 0.344             \\ \hline
crying         & \textbf{0.613}             & reading        & \textbf{0.042}             \\ \hline
driving        & 0.243             & swimming       & 0.189             \\ \hline
exercising     & 0.066             & welding        & \textbf{0.411}             \\ \hline
fishing        & 0.140             & yawning        & 0.157             \\ \hline
\end{tabular}
\end{table}

\subsection{Visual Features}
\subsubsection{Spatial Features}

The Top-1 and the Top-5 Accuracies for the spatial Architectures discussed in section \ref{SpatialFeatures} are shown in Table \ref{MomentsVisual}. The results are comparable to those of the ResNet-50 architecture.

\begin{table}[H]
\begin{center}
\caption{Results on Moments Dataset (Mini Track)}\vspace{2mm}
\begin{tabular}{|c|c|c|}
\hline
Network & Top-1 Accuracy& Top-5 Accuracy\\ \hline
ResNet-50&27.16\% & 51.68\%\\
ResNet-152 & 29.84\% & 54.7\%\\
DenseNet-161 & 25.14\% & 49.12\%\\
VGG-19 with batch normalization & 25.09\% & 50.11\%\\ \hline
\end{tabular}
\end{center}
\label{MomentsVisual}

\end{table}

\begin{figure}[H]
\begin{center}
\includegraphics[scale=0.5]{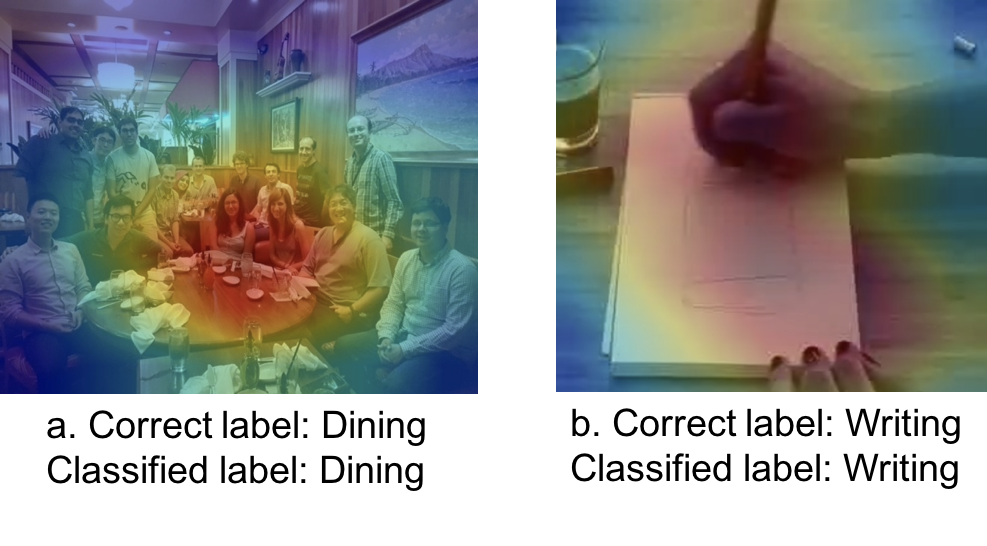}
\caption{Successful Classification with ResNet-50}
\label{SuccessResNet}
\end{center}
\end{figure}

\begin{figure}[H]
\begin{center}
\includegraphics[scale=0.135]{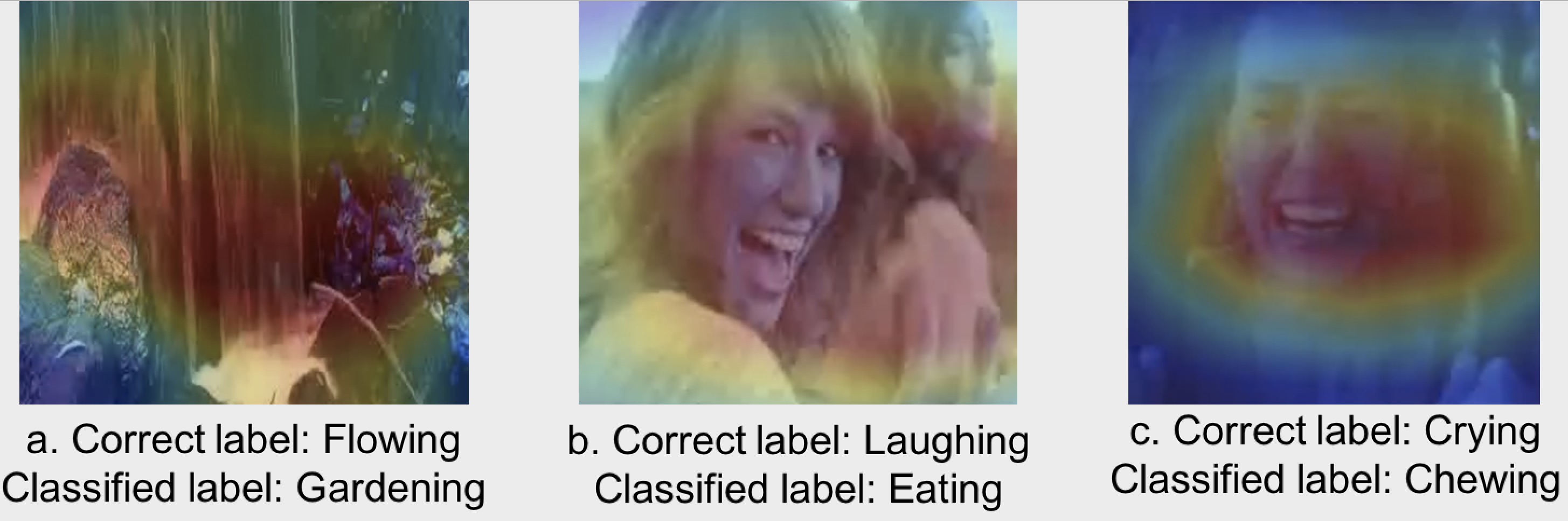}
\caption{Unsuccessful classification with ResNet-50}
\label{FailureResNet}
\end{center}
\end{figure}

The accuracy of spatial features on the Moments of Time Dataset is quite low. In order to analyze the performance of these features, we analyzed the CAM activations performed on the penultimate layer of ResNet-50, the global average pooling layer, to visualize the portions and characteristics of the images the feature was trying to capture\cite{zhou2016learning}. A couple of success and failure cases are depicted in Figure \ref{SuccessResNet} and figure \ref{FailureResNet} respectively.

From these images, we can see that if there exists a strong co-relation between the object detected in the image and the corresponding class, the classification is correct. If not, the classification become unsuccessful.
In image a of \ref{SuccessResNet}, we can see that the architecture places a strong focus on the dining table and the model classifies the class as dining. In image b of Figure \ref{SuccessResNet}, focus is placed on the paper and the writing equipment and the classifier classifies it as writing. For image a of Figure \ref{FailureResNet} the focus is on the rocks below the waterfall. This could be the reason the model classifies it as a gardening class as opposed to flowing. Also, we can see in images b and c of Figure \ref{FailureResNet} that the face of the person is captured, but there are many potentially action based classes that could be associated with the same object such as crying, chewing, eating and laughing. Due to this, the model seems to misclassify in both the instances. We can also assume that the Top 5 accuracy seems fairly higher due to such instances as the true class might actually be present as the next most likely classes as they also depend on concentration of the same object.

\subsubsection{Spatio-Temporal Features}
The ResNext 3D Spatio-Temporal features are primarily used in order to capture the movement of objects over temporal frames. The results for ResNext-3d are as shown in Table \ref{MomentsResNext}.

\begin{table}[H]
\begin{center}
\caption{Results on Moments Dataset (Mini Track)}\vspace{2mm}
\begin{tabular}{|c|c|c|}
\hline
Network & Top-1 Accuracy& Top-5 Accuracy\\ \hline
ResNet-50&27.16\% & 51.68\%\\
ResNet-152 & 29.84\% & 54.7\%\\
ResNext-3D & 45.2\% & 67.4\%\\ \hline
\end{tabular}
\end{center}
\label{MomentsResNext}
\end{table}

\begin{figure}[H]
\begin{center}
\includegraphics[scale=0.5]{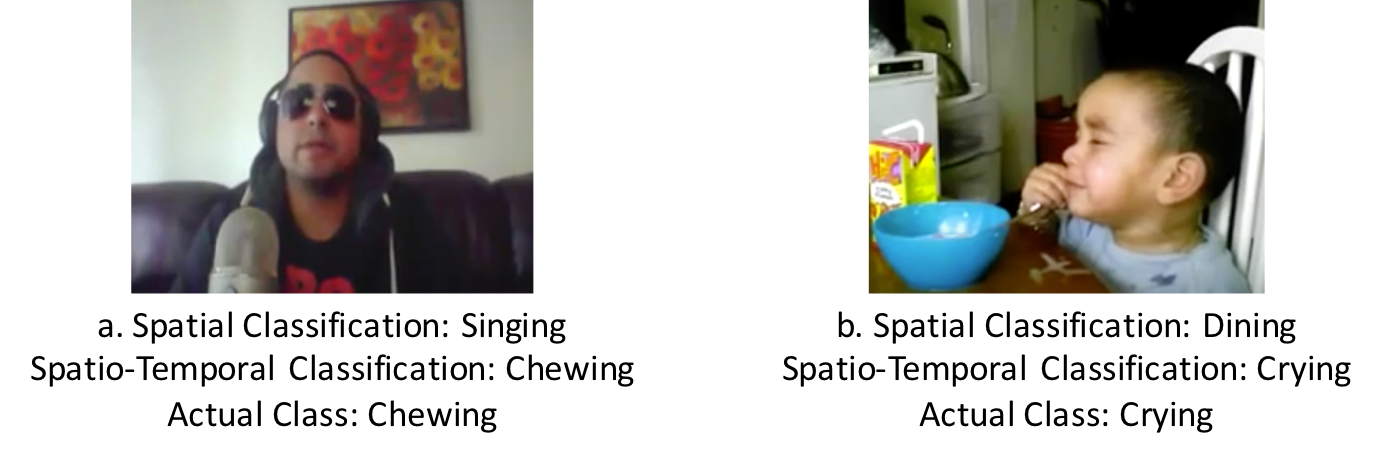}
\caption{Classification comparison with ResNet-50 and ResNext-3D}
\label{HSpatioTemporal}
\end{center}
\end{figure}

\begin{figure}[H]
\begin{center}
\includegraphics[scale=0.5]{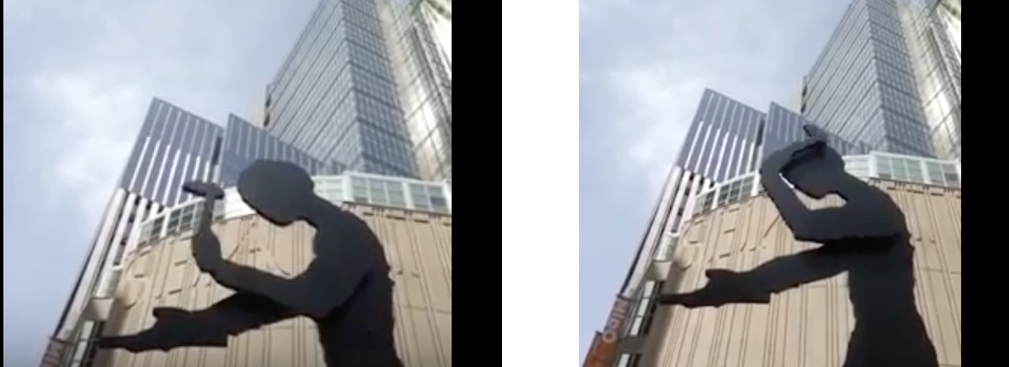}
\caption{Hammering misclassified as Juggling}
\label{Hamering}
\end{center}
\end{figure}

Using ResNext 3D gives a major improvement over using just the spatial features. Several successful examples and their comparison with spatial features are depicted in Figure \ref{HSpatioTemporal} and a few unsuccessful examples are depicted in Figure \ref{Hamering}

The spatial features classify image a of Figure \ref{HSpatioTemporal} as Singing whereas the spatio-temporal features classify this as Chewing. This is probably due to the spatial features paying more attention to the presence of the microphone and the headphones whereas ResNext 3D maybe paying more attention to the chewing activity taking place in the video. Similarly, for image b in Figure \ref{HSpatioTemporal}, the spatial features classify it as Dining, probably due to the presence of the dining table, whereas the spatio temporal features are able to capture the child crying.

Figure \ref{Hamering} depicts an instance where the ResNext 3D features classification is unsuccessful. Here, an act of Hammering is misclassified as Juggling. This may be due to the similarity in actions between both the classes.

\subsection{VisText Analysis}
As described in Section \ref{vistextsection}, we conduct experiments on different Visual Features to generate VisText features. We observe that the Spatial features give a poor accuracy, whereas the 3D-Resnext, which is a feature that capture Spatio-Temporal relations, gives us a very good boost over the initial baseline. 
\begin{table}[H]
    \centering
     \caption{VisText Features Classification results}\vspace{2mm}
    \begin{tabular}{|c|c|} \hline
        Visual Feature & Top-1 Accuracy \\ \hline
        ResNet50 output layer(SVM) & 5.03\% \\
        ResNet50 Penultimate layer(SVM) & 15.34\% \\ 
        3D-ResNext Features (penultimate layer)(SVM) & 54.5\% \\
        3D-ResNext Features (penultimate layer)(MLP) & 56.8\% \\ \hline
    \end{tabular}
    \label{tab:visTextResults}
\end{table}

While training the VisText Feature generation pipeline, we need a metric to evaluate whether the model is performing well or not. Since there is no proper accuracy measure for such a kind of a feature extraction, we qualitatively assess the outcome. Firstly, for all the validation dataset, we extract the 3D-Resnext features, do a forward pass on the trained model and obtain a 300 dimensional vector. We then compute the most similar vector to this vector using gensim. This internally uses a cosine distance to calculate the nearest word to the given 300d vector. Table \ref{vistexttrain} shows some of these results when the model was trained for 35 epochs. 
\begin{table}[H]
\centering
\caption{VisText feature analysis for sample labels} \vspace{2mm}
\label{vistexttrain}
\begin{tabular}{|l|l|l|l|l|}
\hline
Actual Label Word & First Nearest Word & Confidence & Second Nearest Word        & Confidence \\ \hline
chewing           & yawning            & 0.788      & crying                     & 0.682      \\ \hline
juggling          & juggling           & 0.969      & juggle                     & 0.773      \\ \hline
painting          & painting           & 0.895      & tile\_carpentry\_sheetrock & 0.652      \\ \hline
welding           & baking             & 0.945      & cooking                    & 0.676      \\ \hline
yawning           & yawning            & 0.876      & chewing                    & 0.629      \\ \hline
opening           & opening            & 0.692      & closing                    & 0.471      \\ \hline
\end{tabular}
\end{table}
We observe that for classes like \textit{juggling} and \textit{painting} the nearest word to the extracted VisText feature is predicted to be the same with very high confidence. In confusable classes like \textit{opening} and \textit{closing}, our model gives a slightly more confidence to the correct class which confirms our hypothesis. Thus, these features work well and can be used successfully in classification. 
Below are some examples that perform well by VisText, and some that do not perform well. 
\begin{table}[H]
\centering
\caption{Success case analysis for Vis Text feature}\vspace{2mm}
\label{vissuccess}
\begin{tabular}{|l|l|l|l|}
\hline
Actual Label Word & Generated word feature & Classification & Probability  \\ \hline
painting          & painting,carpentry     & painting       & 0.895 \\ \hline
\end{tabular}
\end{table}
It can be seen that in Table \ref{vissuccess} the generated word features nearest neighbours are in fact confusing classes, and due to training a multilayer perceptron to capture these difference, we are able to obtain a higher confidence for the correct class. Thus increasing the label dimensionality from 1 to 300 greatly improves the accuracy of the classification.
\begin{table}[H]
\centering
\caption{Failure Case analysis for Vis Text Feature}\vspace{2mm}
\label{visfailure}
\begin{tabular}{|l|l|l|l|}
\hline
Actual Label Word & Generated word feature & Classification & Probability  \\ \hline
baking            & welding, baking        & welding        & 0.637 \\ \hline
\end{tabular}
\end{table}

As seen in Table \ref{visfailure}, we can observe that although this feature gives the classification as welding, baking is present in the top two. This is a confusing class since in the actual video (present in the final presentation) contains rotating metal plates.

\subsection{Fusion Techniques}
\begin{figure}[H]
\begin{center}
\includegraphics[scale=0.3]{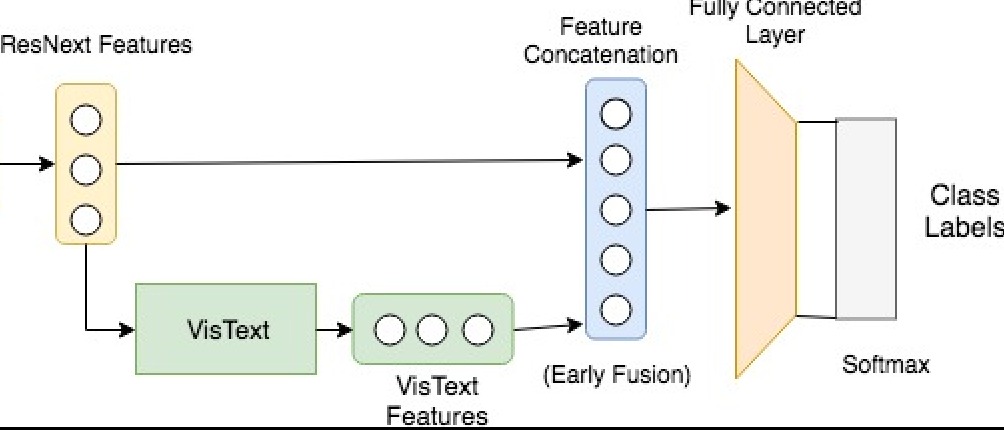}
\caption{ResNext features and VisText Features Hypothesis Early Fusion Model}
\label{fig:early}
\end{center}
\end{figure}

\begin{figure}[H]
\begin{center}
\includegraphics[scale=0.5]{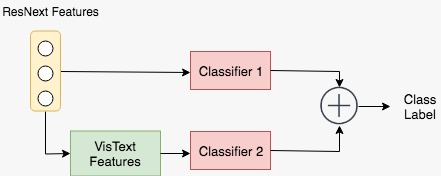}
\caption{ResNext features and VisText Features Hypothesis Late Fusion Model}
\label{fig:late}
\end{center}
\end{figure}

We know that fusion techniques help to boost the classifier accuracies as seen from prior work. Thus, we aim to use the different methods of fusion techniques available and use them with the audio feature. We use Late Fusion as well as Early fusion approach to fuse the classifier information. Reviewing the audio features results, we hypothesized that audio features may not perform well and hence thought that our final architecture would look as below figures \ref{fig:early}, \ref{fig:late}

We observe that for cases of using Logistic Regression as classifier with Late Fusion performs best with ResNext + VisText + Audio features. Note: 60.1 percent is performance when we append zeros for files which doesn't contain audio. Given that around 45 percent of the files don't contain audio appending zeros to those samples during training would create a high bias. Hence, we perform another experiment which gives about 2 percent improvement to \textbf{63.84} where we use only audio files and train a classifier with 3 features and train another classifier with 2 features and then pick the best performance (max of the probability score) for each sample video file. 

\begin{table}[H]
\centering
\caption{Results from various Fusion Techniques - 20 class dataset}\vspace{2mm}
\label{my-label}
\resizebox{\columnwidth}{!}{
\begin{tabular}{|l|l|l|}
\hline
\textbf{Model}                                   & \textbf{Top - 1  Accuracy (\%)} & \textbf{Top - 5 Accuracy (\%)} \\ \hline
Baseline TRN - Moments in Time                   & 31.3                           & 57.4                           \\ \hline
ResNext + VisText (Early Fusion) - (SVM)         & 53.1                           & 79.3                           \\ \hline
ResNext + VisText (Early Fusion) - (MLP)         & 60.0                           & 86.0                           \\ \hline
ResNext + VisText (Late Fusion) - (LR)           & 61.5                           & 87.8                           \\ \hline
ResNext + VisText + Audio  (Early Fusion)  (SVM) & 52.7                           & 77.8                           \\ \hline
ResNext + VisText + Audio  (Early Fusion) - (MLP) & 61.4                           & 86.8                           \\ \hline
ResNext + VisText + Audio (Late Fusion) - (LR)   & 60.1                           & 87.7                           \\ \hline
ResNext + VisText + Only Audio (Late Fusion) - (LR)   & 63.1                           & 89.1                           \\ \hline
ResNext + VisText + Only Audio (Late Fusion) - (Multiple LR)   & \textbf{63.84}                          & \textbf{89.23}                           \\ \hline
\end{tabular}}
\end{table}


\section{Conclusion}
We implemented various state of the art methods in addition to providing a comprehensive analysis of different features namely visual features - spatial and Spatio temporal features, audio features, Textual features - Vistext features. Spatio-Temporal Features provide a richer representation of the video when compared with state-of-the art Spatial Features and VisText Features with the Word2Vec representation trained on a different corpus brings common sense.  into the network and confirms our hypothesis. Audio Features are generally seen to not perform as well as Visual and Textual features which could be attributed to the nature of data as well as sample files with audio content. Thus, we conclude stating that there is a long way to go for a really Intelligent AI.

\section{Future Work}
We have seen visual attention based models can also result in improved accuracy detection scores. Further, we have explored optical flow features but due to the large disk space utilization about 300Gb for 10 classes feature extraction we were unable to perform experiments using that features and we wish to extract optical flow features and then train it using the I3D architecture to observe performance improvements. It would also be interesting to explore as to how handcrafted features such as SURF performs with a temporal model like a RNN or an LSTM. 

\section{Acknowledgement}
We wish to thank Professors Alexander Hauptmann and Florian Metze for their advice. Having access to Bridges computing resources enabled us to work efficiently and produce results on this challenging dataset.

\bibliography{11785_project}

\begin{thebibliography}{19}
\providecommand{\natexlab}[1]{#1}
\providecommand{\url}[1]{\texttt{#1}}
\expandafter\ifx\csname urlstyle\endcsname\relax
  \providecommand{\doi}[1]{doi: #1}\else
  \providecommand{\doi}{doi: \begingroup \urlstyle{rm}\Url}\fi

\bibitem[Aytar et~al.(2016)Aytar, Vondrick, and Torralba]{aytar2016soundnet}
Yusuf Aytar, Carl Vondrick, and Antonio Torralba.
\newblock Soundnet: Learning sound representations from unlabeled video.
\newblock In \emph{Advances in Neural Information Processing Systems}, 2016.

\bibitem[Dong et~al.(2016)Dong, Li, and Snoek]{dong2016word2visualvec}
Jianfeng Dong, Xirong Li, and Cees~GM Snoek.
\newblock Word2visualvec: Cross-media retrieval by visual feature prediction.
\newblock \emph{arXiv preprint arXiv:1604.06838}, 2016.

\bibitem[Fabian Caba~Heilbron \& Niebles(2015)Fabian Caba~Heilbron and
  Niebles]{caba2015activitynet}
Bernard~Ghanem Fabian Caba~Heilbron, Victor~Escorcia and Juan~Carlos Niebles.
\newblock Activitynet: A large-scale video benchmark for human activity
  understanding.
\newblock In \emph{Proceedings of the IEEE Conference on Computer Vision and
  Pattern Recognition}, pp.\  961--970, 2015.

\bibitem[Feichtenhofer et~al.(2016)Feichtenhofer, Pinz, and
  Zisserman]{feichtenhofer2016convolutional}
Christoph Feichtenhofer, Axel Pinz, and AP~Zisserman.
\newblock Convolutional two-stream network fusion for video action recognition.
\newblock 2016.

\bibitem[Hara et~al.(2017)Hara, Kataoka, and Satoh]{hara2017can}
Kensho Hara, Hirokatsu Kataoka, and Yutaka Satoh.
\newblock Can spatiotemporal 3d cnns retrace the history of 2d cnns and
  imagenet?
\newblock \emph{arXiv preprint arXiv:1711.09577}, 2017.

\bibitem[He et~al.(2016)He, Zhang, Ren, and Sun]{7780459}
K.~He, X.~Zhang, S.~Ren, and J.~Sun.
\newblock Deep residual learning for image recognition.
\newblock In \emph{2016 IEEE Conference on Computer Vision and Pattern
  Recognition (CVPR)}, pp.\  770--778, June 2016.
\newblock \doi{10.1109/CVPR.2016.90}.

\bibitem[Kang \& Wildes(2016)Kang and Wildes]{kang2016review}
Soo~Min Kang and Richard~P Wildes.
\newblock Review of action recognition and detection methods.
\newblock \emph{arXiv preprint arXiv:1610.06906}, 2016.

\bibitem[Kumar et~al.(2017)Kumar, Khadkevich, and Fugen]{kumar2017knowledge}
Anurag Kumar, Maksim Khadkevich, and Christian Fugen.
\newblock Knowledge transfer from weakly labeled audio using convolutional
  neural network for sound events and scenes.
\newblock \emph{arXiv preprint arXiv:1711.01369}, 2017.

\bibitem[Monfort et~al.()Monfort, Zhou, Bargal, Yan, Andonian, Ramakrishnan,
  Brown, Fan, Gutfruend, Vondrick, et~al.]{monfortmoments}
Mathew Monfort, Bolei Zhou, Sarah~Adel Bargal, Tom Yan, Alex Andonian, Kandan
  Ramakrishnan, Lisa Brown, Quanfu Fan, Dan Gutfruend, Carl Vondrick, et~al.
\newblock Moments in time dataset: one million videos for event understanding.

\bibitem[Nair \& Hinton(2010)Nair and Hinton]{nair2010rectified}
Vinod Nair and Geoffrey~E Hinton.
\newblock Rectified linear units improve restricted boltzmann machines.
\newblock In \emph{Proceedings of the 27th international conference on machine
  learning (ICML-10)}, pp.\  807--814, 2010.

\bibitem[Ng et~al.(2015)Ng, Hausknecht, Vijayanarasimhan, Vinyals, Monga, and
  Toderici]{ng2015beyond}
Joe Yue-Hei Ng, Matthew Hausknecht, Sudheendra Vijayanarasimhan, Oriol Vinyals,
  Rajat Monga, and George Toderici.
\newblock Beyond short snippets: Deep networks for video classification.
\newblock In \emph{Computer Vision and Pattern Recognition (CVPR), 2015 IEEE
  Conference on}, pp.\  4694--4702. IEEE, 2015.

\bibitem[Shah et~al.(2018)Shah, Kumar, Hauptmann, and Raj]{shah2018closer}
Ankit Shah, Anurag Kumar, Alexander~G Hauptmann, and Bhiksha Raj.
\newblock A closer look at weak label learning for audio events.
\newblock \emph{arXiv preprint arXiv:1804.09288}, 2018.

\bibitem[Simonyan \& Zisserman(2014)Simonyan and Zisserman]{simonyan2014two}
Karen Simonyan and Andrew Zisserman.
\newblock Two-stream convolutional networks for action recognition in videos.
\newblock In \emph{Advances in neural information processing systems}, pp.\
  568--576, 2014.

\bibitem[Venugopalan et~al.(2015)Venugopalan, Rohrbach, Donahue, Mooney,
  Darrell, and Saenko]{venugopalan2015sequence}
Subhashini Venugopalan, Marcus Rohrbach, Jeffrey Donahue, Raymond Mooney,
  Trevor Darrell, and Kate Saenko.
\newblock Sequence to sequence-video to text.
\newblock In \emph{Proceedings of the IEEE international conference on computer
  vision}, pp.\  4534--4542, 2015.

\bibitem[Xie et~al.(2017)Xie, Girshick, Doll{\'a}r, Tu, and
  He]{xie2017aggregated}
Saining Xie, Ross Girshick, Piotr Doll{\'a}r, Zhuowen Tu, and Kaiming He.
\newblock Aggregated residual transformations for deep neural networks.
\newblock In \emph{Computer Vision and Pattern Recognition (CVPR), 2017 IEEE
  Conference on}, pp.\  5987--5995. IEEE, 2017.

\bibitem[Xiong et~al.(2016)Xiong, Wang, Wang, Zhang, Song, Li, Lin, Qiao,
  Van~Gool, and Tang]{xiong2016cuhk}
Yuanjun Xiong, Limin Wang, Zhe Wang, Bowen Zhang, Hang Song, Wei Li, Dahua Lin,
  Yu~Qiao, Luc Van~Gool, and Xiaoou Tang.
\newblock Cuhk \& ethz \& siat submission to activitynet challenge 2016.
\newblock \emph{arXiv preprint arXiv:1608.00797}, 2016.

\bibitem[Xu \& Zhang(2017)Xu and Zhang]{xu2017attngan}
Tao Xu and et~al. Zhang.
\newblock Attngan: Fine-grained text to image generation with attentional
  generative adversarial networks.
\newblock \emph{arXiv preprint arXiv:1711.10485}, 2017.

\bibitem[Zhou et~al.(2016)Zhou, Khosla, Lapedriza, Oliva, and
  Torralba]{zhou2016learning}
Bolei Zhou, Aditya Khosla, Agata Lapedriza, Aude Oliva, and Antonio Torralba.
\newblock Learning deep features for discriminative localization.
\newblock In \emph{Computer Vision and Pattern Recognition (CVPR), 2016 IEEE
  Conference on}, pp.\  2921--2929. IEEE, 2016.

\bibitem[Zhou et~al.(2017)Zhou, Andonian, and Torralba]{zhou2017temporal}
Bolei Zhou, Alex Andonian, and Antonio Torralba.
\newblock Temporal relational reasoning in videos.
\newblock \emph{arXiv preprint arXiv:1711.08496}, 2017.

\end{thebibliography}
\bibliographystyle{11785_project}

\end{document}